\documentclass[final]{cvpr}

\usepackage{times}
\usepackage{epsfig}
\usepackage{graphicx}
\usepackage{amsmath}
\usepackage{amssymb}
\usepackage{multirow}
\usepackage[table,xcdraw]{xcolor}
\usepackage{tabularx}
\usepackage{booktabs}
\usepackage{capt-of}
\usepackage{color}

\newcommand{\nelson}[1]{}
\newcommand{\yasu}[1]{}
\newcommand{\kai}[1]{}
\newcommand{\chin}[1]{}
\newcommand{\greg}[1]{}
\newcommand{\hang}[1]{}

\newcommand{\mysubsubsection}[1]{\vspace{0.15cm} \noindent {\bf #1}:}

\usepackage{xcolor}
\usepackage{array}
\usepackage{varwidth}

\definecolor{darkgreen}{RGB}{34,139,34}

\usepackage[pagebackref=true,breaklinks=true,colorlinks,bookmarks=false]{hyperref}
\usepackage{comment}

\def\cvprPaperID{1834} 
\def\confYear{CVPR 2021}

\begin{document}

\title{House-GAN++: Generative Adversarial Layout Refinement Networks}


\author{Nelson Nauata\\
Simon Fraser University\\
{\tt\small nnauata@sfu.ca}
\and
Sepidehsadat Hosseini\\
Simon Fraser University\\
{\tt\small sepidh@sfu.ca}

\and
Kai-Hung Chang\\
Autodesk Research\\
{\tt\small kai-hung.chang@autodesk.com}

\and
Hang Chu\\
Autodesk Research\\
{\tt\small hang.chu@autodesk.com}

\and
Chin-Yi Cheng\\
Autodesk Research\\
{\tt\small chin-yi.cheng@autodesk.com}

\and
Yasutaka Furukawa\\
Simon Fraser University\\
{\tt\small furukawa@sfu.ca}
}

\twocolumn[{
\maketitle
\vspace{1em}
\centerline{
\includegraphics[width=\linewidth]{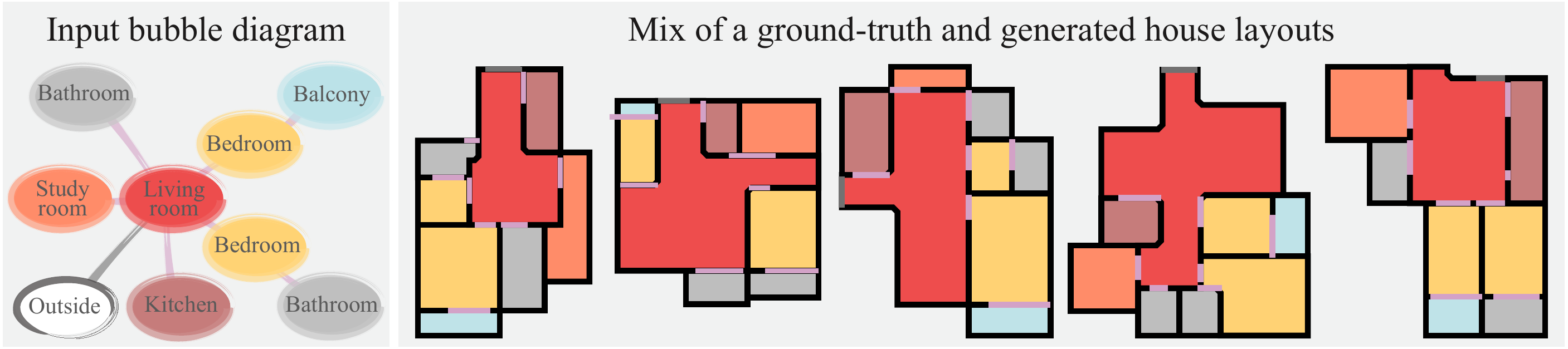}
}
\captionof{figure}{The paper makes a breakthrough in the task of automated house layout generation. The right shows the mix of a ground-truth design made by an architect and our generated samples, based on the input bubble-diagram. Can you tell which one is the ground-truth? See the end of the caption for the answer.
The paper proposes a novel generative adversarial layout refinement network, whose generator is trained to repeatedly apply and refine the design towards perfection. 
(The second sample from the right is the ground-truth.)
%
}
\label{fig:teaser}
\vspace{1em}
}]

\begin{abstract}
This paper proposes a novel generative adversarial layout refinement network for automated floorplan generation.
%
Our architecture is an integration of a graph-constrained relational GAN
and a conditional GAN, where a previously generated layout becomes the next input constraint, enabling iterative refinement.
%
A surprising discovery of our research is that a simple non-iterative training process, dubbed component-wise GT-conditioning, is  effective in learning such a generator.
%
The iterative generator also creates a new opportunity in further improving a metric of choice via meta-optimization techniques 
by controlling when to pass which input constraints during iterative layout refinement.
%
%
%
Our qualitative and quantitative evaluation based on the three standard metrics demonstrate that the proposed system makes significant improvements over the current state-of-the-art, even competitive against the ground-truth floorplans, designed by professional architects.
We will share our code, model, and data.

\end{abstract}
\begin{figure*}[!t]
     \centering
     \includegraphics[width=\linewidth]{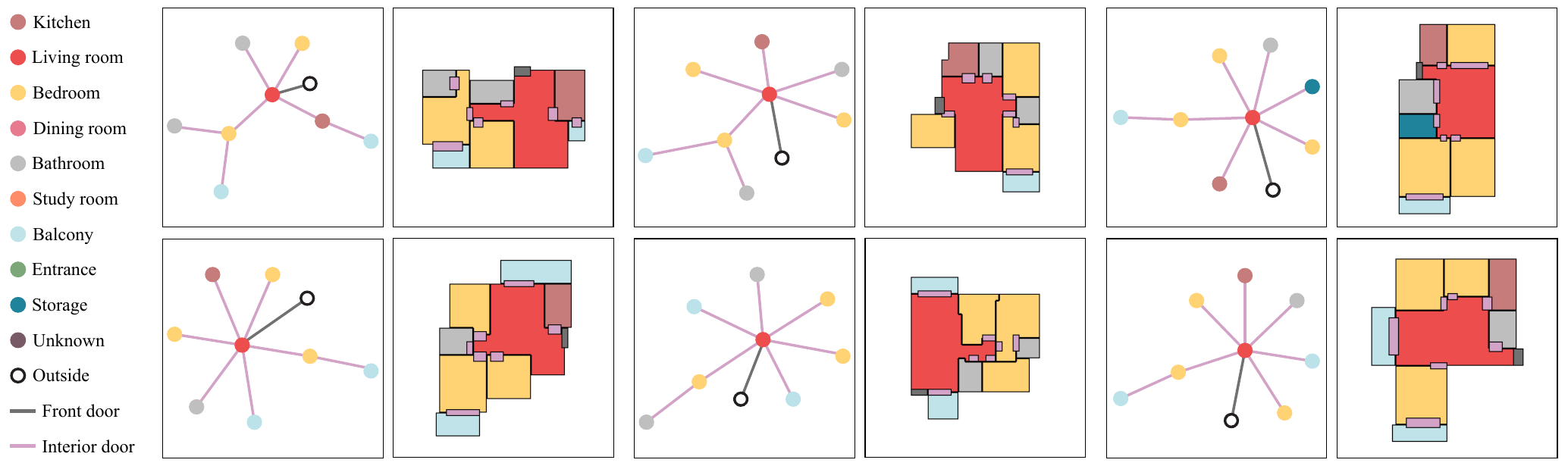}
     \caption{We have parsed the RPLAN dataset~\cite{wu2019data} to prepare 60k house layouts with the corresponding bubble-diagrams.
     Our task is to generate a realistic and diverse set of floorplans that are compatible with the input bubble-diagram.
     } 
    \label{figure:groundtruth}
\end{figure*} 

\section{Introduction}

House design is a time-consuming iterative process, requiring multiple rounds of refinements.
An architect sketches out a design, evaluates, adjusts, and repeats the cycles until being satisfied with a design
within a given time budget.
Unfortunately, designing an effective floorplan is possible only by professional architects, where a small fraction of buildings (less than 10\% in North America) employ a design professional for custom design due to the cost. Automatic floorplan generation will have tremendous impacts on the trillion-dollar real-estate/construction industries.

Floorplan generation has been an active area of research. Early methods formulated it as iterative optimization~\cite{merrell2010computer,martin2006procedural}. The surge of deep neural networks has made a breakthrough, where state-of-the-art algorithms~\cite{nauata2020house,chen2020intelligent} utilize Generative Adversarial Networks (GANs)~\cite{goodfellow2014generative}.
%
GAN is a one-shot generation process, converting a noise vector into a sample. A conditional GAN could take a design in progress and produce a refined version iteratively~\cite{pathak2016context}. However, there does not exist a database of incomplete floorplans during design iterations, making such a training infeasible.

This paper proposes a novel generative adversarial layout refinement network, where the generator is trained to generate a floorplan as a graph of segmentation masks by iteratively refining its design (See Fig.~\ref{fig:teaser}).
We borrow the problem setup from the state-of-the-art system House-GAN~\cite{nauata2020house}, while making a few extensions towards a more challenging production-level task (i.e., handling non-rectangular room shapes, generating doors/entrances, and using a functional graph instead of an adjacency graph).

The technical contribution of this paper is three-fold: an architecture, a training algorithm, and a test-time meta-optimization algorithm. First, the architecture is an integration of a graph-constrained relational GAN (i.e., based on the current state-of-the-art~\cite{nauata2020house}) and a conditional GAN~\cite{pathak2016context}, where a previously generated model becomes the next input constraint, enabling iterative layout refinement. Second, a surprising discovery of our research is that a computationally affordable non-iterative training process, dubbed component-wise GT-conditioning, is effective in learning an iterative generator, where a ground-truth segmentation is passed to each component as a condition at a random probability.
Third, our framework creates a new opportunity in further optimizing a metric of choice via meta-optimization, such as Bayesian optimization by controlling when to pass which constraints.

We have used the RPLAN dataset~\cite{wu2019data}, which offers
60k vector-graphics floorplans designed by professional architects.
%
Qualitative and quantitative evaluations based on the three standard metrics (i.e., realism, diversity, and compatibility) in the literature demonstrate that the proposed system outperforms the current-state-of-the-art by a large margin. The system is even competitive against the ground-truth floorplans, based on user studies judged by professional architects. We will share our code, model, and data.


\begin{figure*}[!t]
     \centering
     \includegraphics[width=\linewidth]{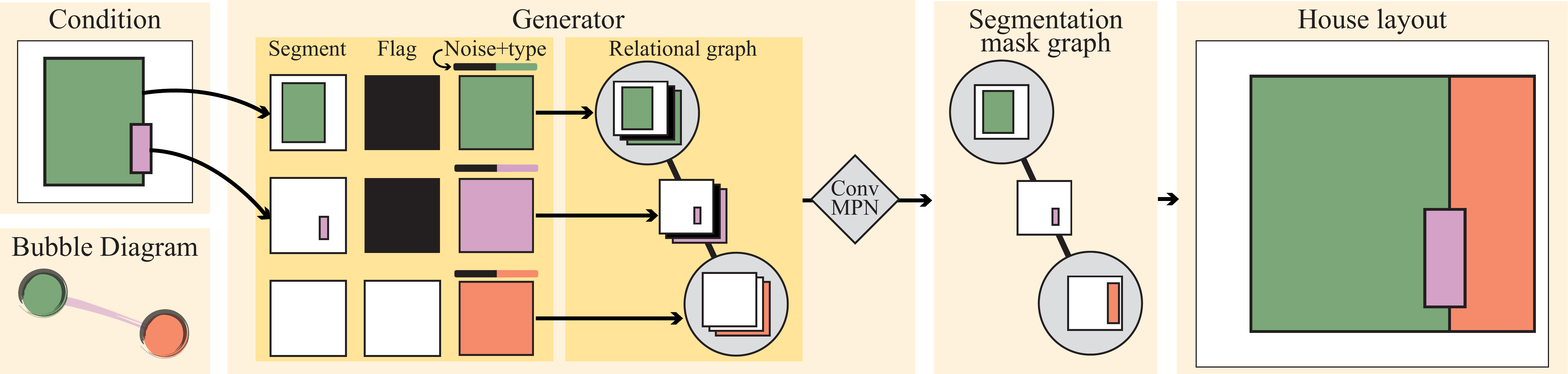}
     \caption{Our architecture is built on top of a relational GAN from the state-of-the-art system~\cite{nauata2020house}.
     An additional 2D segmentation mask for each room/door can be specified as an input condition, enabling iterative design refinement.
     }
    \label{figure:generator_architecture}
\end{figure*} 

\section{Related Work}



This section reviews generative models for structured data such as objects, 3D buildings, or floorplans.

\mysubsubsection{Traditional methods}
Classical algorithms formulate optimization or design hand-crafted rules for structured data generation.
%
Procedural modeling is an early successful approach for 3D building models, where shape-grammars define iterative generation processes~\cite{muller2006procedural,bao2013generating}. 
Integer programming is formulated to generate an arrangement of tiles~\cite{peng2014computing}.
Optimization has also proven effective for game-level design~\cite{ma2014game,hendrikx2013procedural}.
For floorplans specifically, Bayesian network is trained to learn distributions of architectural components for stochastic sample generation~\cite{merrell2010computer}.

\mysubsubsection{Neural approach (one-step generation)}
With the surge of deep learning, researchers started to train deep neural networks (DNNs) to learn a single-step model generation.
Convolutional neural networks are trained to add objects into a room one by one for indoor scene generation~\cite{wang2018deep,ritchie2019fast}.
A scene graph generation in addition to the object placement are learned step by step via DNNs~\cite{wang2019planit}.
Neural turtle graphics learns RNNs to encode incoming paths and generate outgoing edges at one node~\cite{chu2019neural}. An iterative algorithm uses the RNN modules repeatedly to generate road layouts. 
While being a reconstruction task
as opposed to a generative one, 
RoadTracer learns a CNN that takes the current partial network then decides on a next action for reconstructing a road network from a satellite image~\cite{bastani2018roadtracer}.

\mysubsubsection{Neural approach (joint generation)}
Joint generation of multiple components in an entire structured model is more challenging. Room boundaries as a raster image is estimated from a house foot-print by a CNN, followed by a series of heuristics to generate a vectorized floorplan~\cite{wu2019data}.
Image generation has been an active area of research, where realistic object arrangements are learned via variational auto encoder~\cite{jyothi2019layoutvae}, a differentiable renderer~\cite{li2019layoutgan}, or graph convolutional networks~\cite{johnson2018image,ashual2019specifying}.
%
Graph generation is modeled as a sequence of graph editing operations, whose entire process is learned by recurrent neural networks~\cite{you2018graphrnn} or graph recurrent attention networks~\cite{liao2019efficient}. Natural language generation is another related area of research, where an auto-regressive model is trained by masking words from ground-truth sentences in the input~\cite{liao2020probabilistically}.
%
The adversarial training is also employed for the same language task~\cite{fedus2018maskgan} in a seq2seq architecture. 
This paper takes the idea of masking and adversarial training to the task of 2D layout generation, where our architecture jointly estimates segmentation masks for all architectural components (as opposed to a recurrent formulation) and is capable of iteratively refining a design.




\mysubsubsection{Neural approach (joint generation w/ refinement)}
On the iterative sample refinement, GANHopper is the closest to ours in spirit~\cite{lira2020ganhopper}. The key difference is that our system produces a structured model (as opposed to a raster image) and the training process is non-sequential.

\section{Problem Formulation} \label{sect:problem}

In a standard design process, architects incorporate constraints from clients into a sketch called a bubble-diagram, then convert it into a floorplan through design explorations and iterations.
We borrow the problem setup from the state-of-the-art system House-GAN~\cite{nauata2020house}, while adding a few extensions towards the production-level task.

\mysubsubsection{Input}
The input to the system is a bubble-diagram, which is represented as a graph where a node encodes a room with its room type (See Fig.~\ref{figure:groundtruth})~\footnote{Room types are: ``living room'', ``kitchen'', ``bedroom'', ``balcony'', ``entrance'', ``dining room'', ``study room'', ``storage'', ``unkown'', or ``outside''. Note that ``outside'' is not an actual room and we define for the convenience of defining a ``front door'' as an edge.}. In the original problem, an edge encoded spatial adjacency of two rooms. In this work, an edge encodes a functional connection (``interior door'' or ``front door'') as in real architect's sketches. An ``interior door'' is a connection between two rooms, and a ``front door'' is a connection between a room and the outside area.

\mysubsubsection{Output}
The output of the system is a segmentation mask for each room and door.
%
We also utilize an off-the-shelf floorplan vectorization algorithm~\cite{chen2019floor} to convert the layout into a vector floorplan image, where a room is represented by an axis-aligned closed-polygon, adjacent rooms share the walls with the common line segments, and a door is represented as a line-segment on a wall.

\mysubsubsection{Metrics} Realism, diversity, and compatibility are the three metrics, evaluating the performance as in the prior work~\cite{nauata2020house}.
Roughly speaking, realism is an average user rating, diversity is the Fréchet Inception Distance (FID)~\cite{heusel2017gans}, and the compatibility is the graph edit distance (GED)~\cite{abu2015exact} between the input bubble-diagram and the one constructed from the output layout.
To account for the addition of doors,
we adjusted the metric definitions slightly, whose details are referred to the supplementary document. We describe the details of the user study in Sect.~\ref{sect:results}.




\begin{figure*}[!t]
     \centering
     \includegraphics[width=\linewidth]{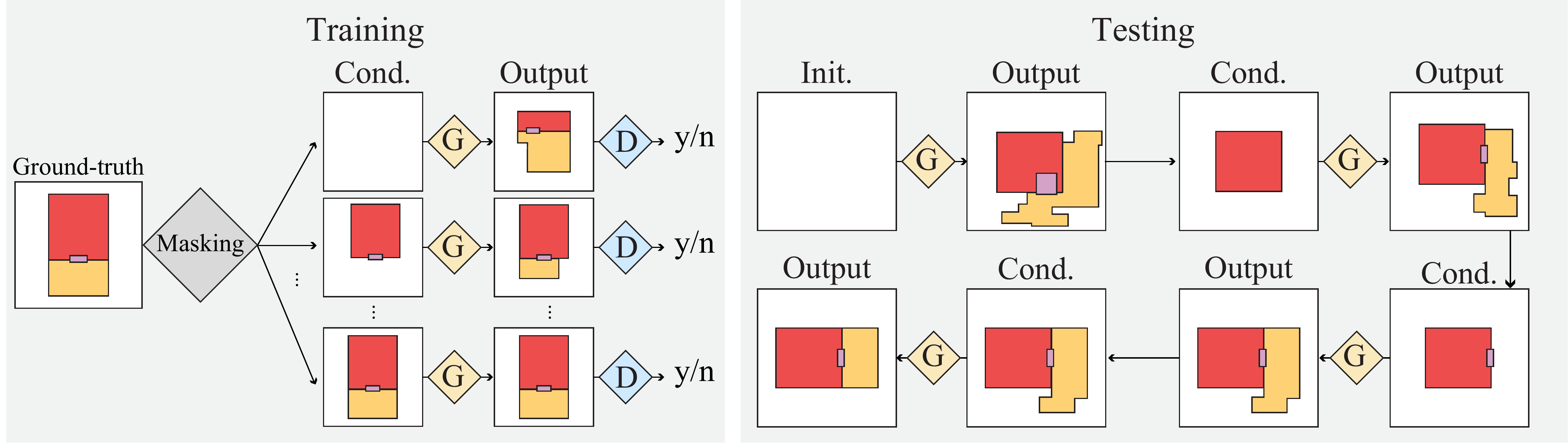}
     \caption{
     During training, we specify a GT segmentation mask for each room/door with a 50\% chance. The generator needs to learn the task of inpainting missing components when many GT masks are given, or the task of generating a complete design when few masks are given.
     %
     During testing, previously generated layouts are passed to the generator as potential input constraints, enabling iterative design refinement.}
    \label{figure:training_testing_formulation}
\end{figure*} 

\section{Technical Innovations}

Learning an iterative refinement process is non-trivial due to the lack of databases containing step-by-step design iterations.
Our architecture is a straightforward integration of a relational GAN~\cite{nauata2020house} and a conditional GAN~\cite{pathak2016context}. A surprising discovery of our research is that a simple non-iterative training procedure is effective in training such an iterative generator. The iterative refinement capability opens up a new opportunity in further improving the metric of choice by meta-optimizing the refinement scheme at test-time. The section explains the architecture, the training algorithm, and the meta-optimization algorithm.

\subsection{Architecture}\label{architecture}
Following House-GAN~\cite{nauata2020house}, our network backbone is a convolutional message passing network (Conv-MPN~\cite{zhang2020conv}), whose relational graph structure is defined by the bubble-diagram (See Fig.~\ref{figure:generator_architecture}). There are three key differences in our architecture: 1) Edges in addition to nodes carry features for the generation of doors; 2) Each node/edge takes a 2D segmentation mask as an additional input constraint with an associated new loss; and 3) Conv-MPN feature pooling~\cite{zhang2020conv} is reformulated to allow feature-exchange between nodes and edges. We now explain the three differences in detail.

\mysubsubsection{Edge features}
House-GAN has a 10-d one-hot vector, which encodes a room type and initializes a node feature vector. In our work, doors have two types, and we extend the one-hot vector to 12-d
over the mix of 10 room types and 2 door types. With the common type vector, door generation from an edge becomes the same as room generation from a node,
except the pooling mechanism in the convolutional message passing as detailed in the following paragraph.


\mysubsubsection{Mask condition}
House-GAN initializes each node with a noise vector and a room-type, which is transformed into a $8\times 8\times 16$ feature volume. Our relational generator takes in an additional $64\times 64\times 2$ condition image for each node/edge. The first channel provides the segmentation mask, which we expect the generator to learn to keep unchanged. The second channel becomes 1 for every pixel when the segmentation mask is specified, otherwise 0.
We use a 3-layer CNN to convert the condition image to $8\times8\times 16$, which is concatenated to the original feature.
When the segmentation mask is specified, we enforce an L1-loss between the condition image $\hat{\mathbf{m}}_{i}$ and the generated mask $\mathbf{m}_i$. $i$ is the index of a node or an edge. Precisely, let $\mathcal{I}$ denote the set of node/edge indices where the condition masks are specified. The loss function is defined as follows, where $L_{org}$ is the original generator adversarial loss~\cite{gulrajani2017improved} in House-GAN and $\lambda$ is $1000$:
\begin{equation}
    L= L_{org}+\lambda \frac{1}{|\mathcal{I}|}\sum_{i\in \mathcal{I}} \mathopen| \mathbf{m}_{i}-\hat{\mathbf{m}}_{i}\mathclose|.
\end{equation}





\mysubsubsection{Conv-MPN pooling}
Conv-MPN message passing is defined based on the connectivity of the relational graph~\cite{zhang2020conv}:
\begin{equation}
    \mathbf{g}_r \leftarrow \mbox{CNN}\left[\mathbf{g}_r\quad ;\quad \underset{s\in \mathcal{N}(r)}{\mbox{Pool}} \mathbf{g}_s \quad ; \quad \underset{s \in \overline{\mathcal{N}}(r)}{\mbox{Pool}} \mathbf{g}_s \right].
\end{equation}
$\mathbf{g}_r$ denotes the feature volume for a component $r$. $\mathcal{N}(r)$ and $\overline{\mathcal{N}}(r)$ denote the neighbors and its complement.
We redefine $\mathcal{N}(r)$ to allow feature exchange between rooms and doors.
First, a room is a neighbor of its connected rooms and the doors in-between. Second, a door is a neighbor of its incidental two rooms. The rest of the architecture is the same as House-GAN, except that we upsample the feature volume three times instead of twice, where the final segmentation mask is produced in the resolution of $64\times 64$.

\subsection{Component-wise GT-conditional training}\label{training}
Our training strategy is similar in spirit to ``masking'' in NLP~\cite{fedus2018maskgan} (See Fig.~\ref{figure:training_testing_formulation}). In each training step, we pick a ground-truth layout at random, initialize a relational graph with its bubble-diagram, then specify a ground-truth segmentation mask as the input condition for each room/door with a 50\% chance.
This strategy forces the generator to either create a whole layout when few GT conditions are specified, or simply inpaint the missing components when many GT conditions are given.
Surprisingly, this simple training process allows the generator to iteratively improve a design by passing the previously generated layout as the input condition.
\subsection{Meta-optimizing iterative refinement scheme}\label{testing}
Iterative layout refinement starts by running a generator without input constraints.
From the second iteration, 
we have an option of specifying the previously generated room/edge masks.
What is the best strategy? Is it the best to always pass all the constraints? 
We propose three strategies that control when to pass which constraints based on: 1) fixed heuristics; 2) static node/edge properties; or 3) dynamic layout information.
In the latter two cases, we parameterize the space of strategies and employ a meta-optimization algorithm to seek for good solutions subject to either our diversity or compatibility metrics. In all strategies, we run the generator 10 times to refine a layout.


\mysubsubsection{Fixed heuristics}
The first heuristic is to pass the segmentation mask for each room/door at a 50\% chance at every iteration ($\mbox{Ours}_{heur}^{50\%}$). The second heuristic is to always specify all the conditions
 ($\mbox{Ours}_{heur}^{100\%}$).


{\vspace{0.15cm} \noindent {\bf Static scheme} ($\mbox{Ours}_{static}$):}
Our static scheme is based on the observation that architects start by designing certain spaces first (e.g., a living room). We parameterize the scheme by a 12-d vector $\{V_i\}$, where $i$ is an index over 10 room types and 2 door types. Each element takes a value in the range $[1, 10]$. Suppose $V_3$ is 4. In this scheme, the previous mask is specified as the input condition for the $3_{rd}$ room (or door) type  after the $4_{th}$ iteration. FID (i.e., diversity) score is the target metric for the optimization.


{\vspace{0.15cm} \noindent {\bf Dynamic scheme} ($\mbox{Ours}_{dyna}$):}
Another intuitive strategy is to control the conditioning based on the compatibility of the current model with the input bubble-diagram.
%
We parameterize the scheme by two 12-d vectors $\{T_i\}$ and $\{U_i\}$, where $i$ is again the index over the room/door type.
Suppose $T_3$ is 4, and $U_3$ is 7. In this scheme, the previously generated mask is specified after the $4_{th}$ (resp. $7_{th}$) iteration if the current room (door) is compatible (resp. incompatible) for the $3_{rd}$ room type. FID (i.e., diversity) or a graph edit distance (i.e., compatibility) is the target metric.
\begingroup
\renewcommand{\arraystretch}{1.1}
\begin{table*}[!t]
\caption{The main quantitative evaluations. Realism is measured by a user study with amateurs and professional architects. Diversity is measured by the FID scores. Compatibility is measured by the graph edit distance. $(\uparrow)$ and $(\downarrow)$ indicate the-higher-the-better and the-lower-the-better metrics, respectively.
The \textcolor{cyan}{cyan}, \textcolor{orange}{orange}, and \textcolor{magenta}{magenta} colors indicate the first, the second, and third best results, respectively.}
\label{tab:main_results}
\centering
\setlength{\tabcolsep}{3.5pt}
\begin{tabular}{p{2.6cm}ccccccccc}
\toprule
    & \multicolumn{1}{c}{Realism $(\uparrow)$} & \multicolumn{4}{c}{Diversity $(\downarrow)$} & \multicolumn{4}{c}{Compatibility $(\downarrow)$} \\
    \cmidrule(lr){2-2}\cmidrule(lr){3-6}\cmidrule(lr){7-10} Model &
    8 & 5 & 6 & 7 & 8 & 5 & 6 & 7 & 8 \\
    \midrule
    
    Ashual \etal ~\cite{ashual2019specifying} & \textcolor{magenta}{-0.7} & \textcolor{magenta}{120.6\textpm0.5} & \textcolor{magenta}{172.5\textpm0.2} & \textcolor{magenta}{162.1\textpm0.4} & \textcolor{magenta}{183.0\textpm0.4} & \textcolor{magenta}{7.5\textpm0.0} & 9.2\textpm0.0 & \textcolor{magenta}{10.0\textpm0.0} & 11.8\textpm0.0\\
    Johnson \etal ~\cite{johnson2018image} & \textcolor{magenta}{-0.7} & 167.2\textpm0.3 & 168.4\textpm0.4 & 186.0\textpm0.4 & 186.0\textpm0.4 & 7.7\textpm0.0 & \textcolor{magenta}{6.5\textpm0.0} & 10.2\textpm0.0 & \textcolor{magenta}{11.3\textpm0.1}\\
    House-GAN~\cite{nauata2020house} & \textcolor{orange}{0.0} &  \textcolor{orange}{37.5\textpm1.1} & \textcolor{orange}{41.0\textpm0.6} & \textcolor{orange}{32.9\textpm1.2} & \textcolor{orange}{66.4\textpm1.7} & \textcolor{orange}{2.5\textpm0.1} & \textcolor{orange}{2.4\textpm0.1} & \textcolor{orange}{3.2\textpm0.0} & \textcolor{orange}{5.3\textpm0.0}\\
    $\mbox{Ours}_{heur}^{50\%}$ & \textcolor{cyan}{0.2} & \textcolor{cyan}{30.4\textpm4.4} & \textcolor{cyan}{37.6\textpm3.0} & \textcolor{cyan}{27.3\textpm4.9} & \textcolor{cyan}{32.9\textpm4.9} & \textcolor{cyan}{1.9\textpm0.3} & \textcolor{cyan}{2.2\textpm0.3} & \textcolor{cyan}{2.4\textpm0.3} & \textcolor{cyan}{3.9\textpm0.5}\\

    \bottomrule
\end{tabular}
\end{table*}
\endgroup

\begingroup
\renewcommand{\arraystretch}{1.4}
\begin{table}[!t]
\caption{Refinement scheme optimization. The diversity and the compatibility metrics are shown for House-GAN and five variants of our method.
Meta optimization is used to optimize the refinement scheme in the last three rows. ``Target'' column indicates the target metric of the scheme-optimization, where ``Divers.'' and ``Compat.'' indicate FID and the graph edit distance metrics.
}
\label{tab:ablation}
\centering
\setlength{\tabcolsep}{3pt}
\begin{tabular}{l|ccc}
\toprule
    Model & Target & \multicolumn{1}{c}{Divers. $(\downarrow)$} & \multicolumn{1}{c}{Compat. $(\downarrow)$} \\
    \midrule
    House-GAN &N/A& 66.4\textpm1.7 & 5.3\textpm0.0\\
    $\mbox{Ours}_{heur}^{50\%}$ &N/A& 32.9\textpm4.9 & 3.9\textpm0.5\\
    $\mbox{Ours}_{heur}^{100\%}$ &N/A& \textcolor{magenta}{31.2\textpm0.6} & \textcolor{magenta}{3.7\textpm0.1}\\
    $\mbox{Ours}_{static}$ &Divers.&  \textcolor{orange}{27.3\textpm1.1} & \textcolor{orange}{3.2\textpm0.1}\\
    
    $\mbox{Ours}_{dyna}$ &Divers.& \textcolor{cyan}{24.7\textpm0.8} & \textcolor{magenta}{3.7\textpm0.0}\\
    
    $\mbox{Ours}_{dyna}$ &Compat.& 35.7\textpm1.3 & \textcolor{cyan}{2.6\textpm0.0}\\

    \bottomrule
\end{tabular}
\end{table}
\endgroup

\begin{figure*}[!t]
     \centering
     \includegraphics[width=0.92\linewidth]{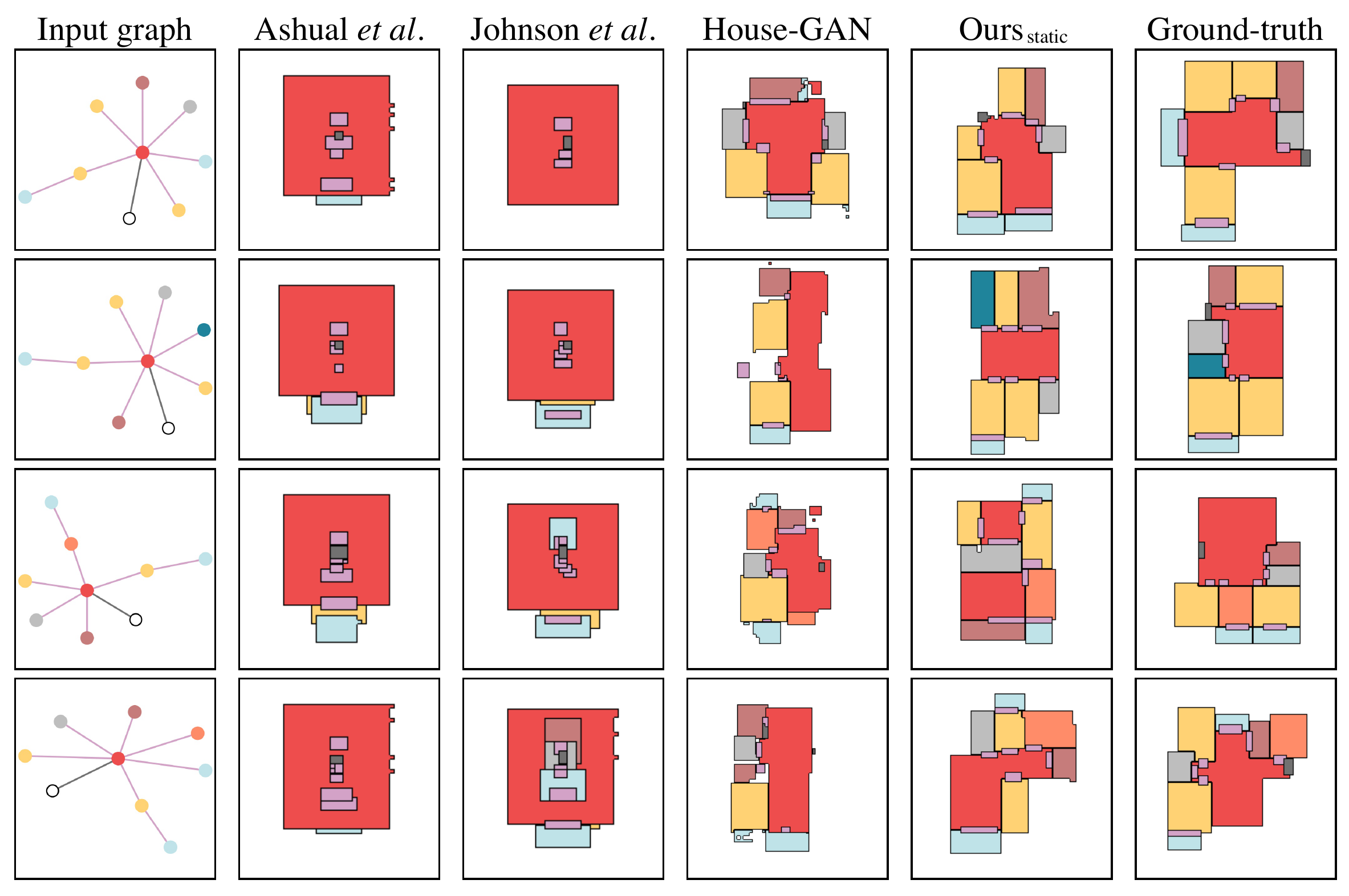}
     \caption{Realism evaluation. One generated layout is shown for each input bubble diagram.
     %
    }
    \label{figure:realism}
\end{figure*} 

\begin{figure*}[!tb]
     \centering
     \includegraphics[width=0.92\linewidth]{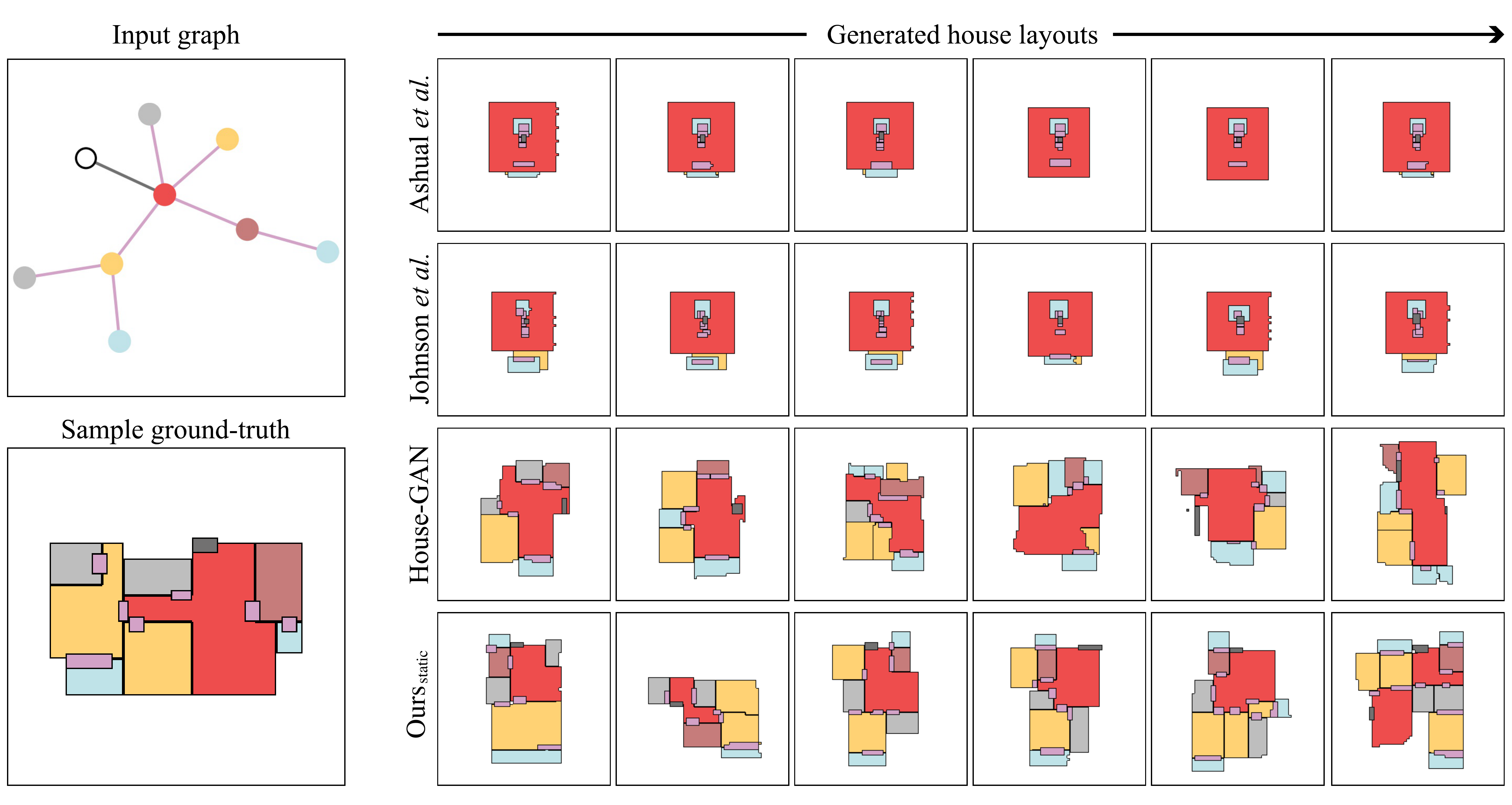}
     \caption{Diversity evaluation. Given an input bubble-diagram, we show six samples generated by each method. 
     }
    \label{figure:diversity}
\end{figure*} 

\begin{figure*}[!tb]
     \centering
     \includegraphics[width=0.92\linewidth]{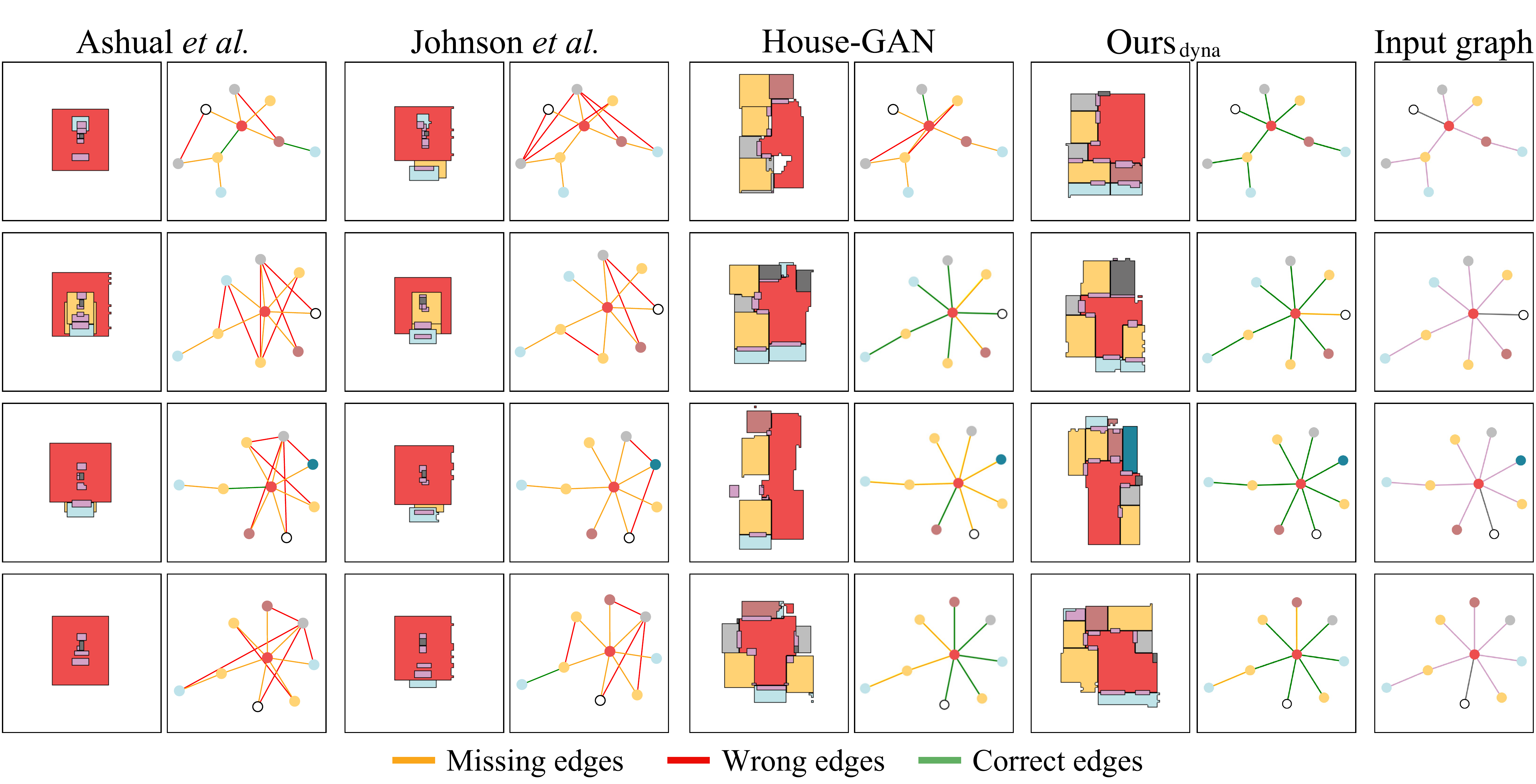}
     \caption{Compatibility evaluation. The figure shows the inconsistency between the input bubble diagram and the ones constructed by the output layouts.
     The \textcolor{orange}{orange}, \textcolor{red}{red} and \textcolor{darkgreen}{green} colors indicate if an edge is missing, wrongly predicted, or correctly predicted.
     }
    \label{figure:compatibility}
\end{figure*} 

\begin{figure*}[!tb]
     \centering
     \includegraphics[width=0.92\linewidth]{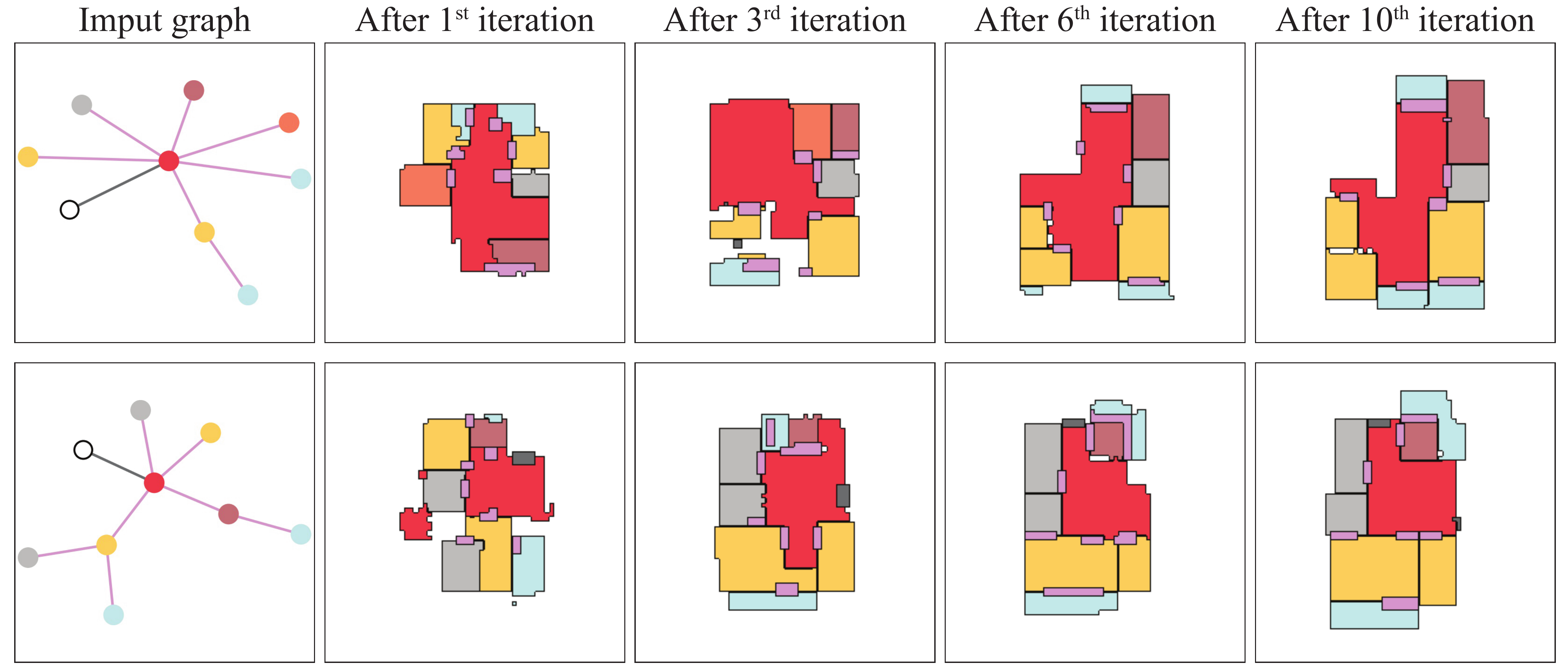}
     \caption{Iterative layout refinement. Each row shows the input bubble-diagram and the iterative refinement process over the 10 iterations.
     %
     }
    \label{figure:iterations}
\end{figure*} 

\begin{figure*}[!tb]
     \centering
     \includegraphics[width=0.92\linewidth]{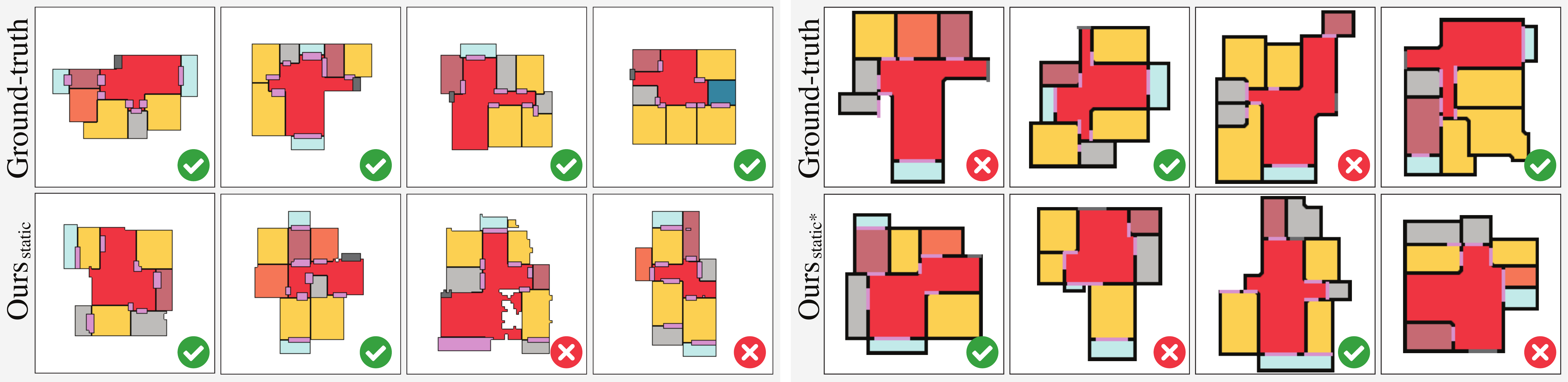}
     \caption{
     Representative user study results. The left is the raw segmentation visualization and the right is the vector-floorplan visualization.
     Each column is a pair presented to the subjects. A green mark denotes the preferred design. Two green marks denote a tie.
     }
    \label{figure:user_study}
\end{figure*}

\begin{figure}[!tb]
     \centering
    \includegraphics[width=\columnwidth]{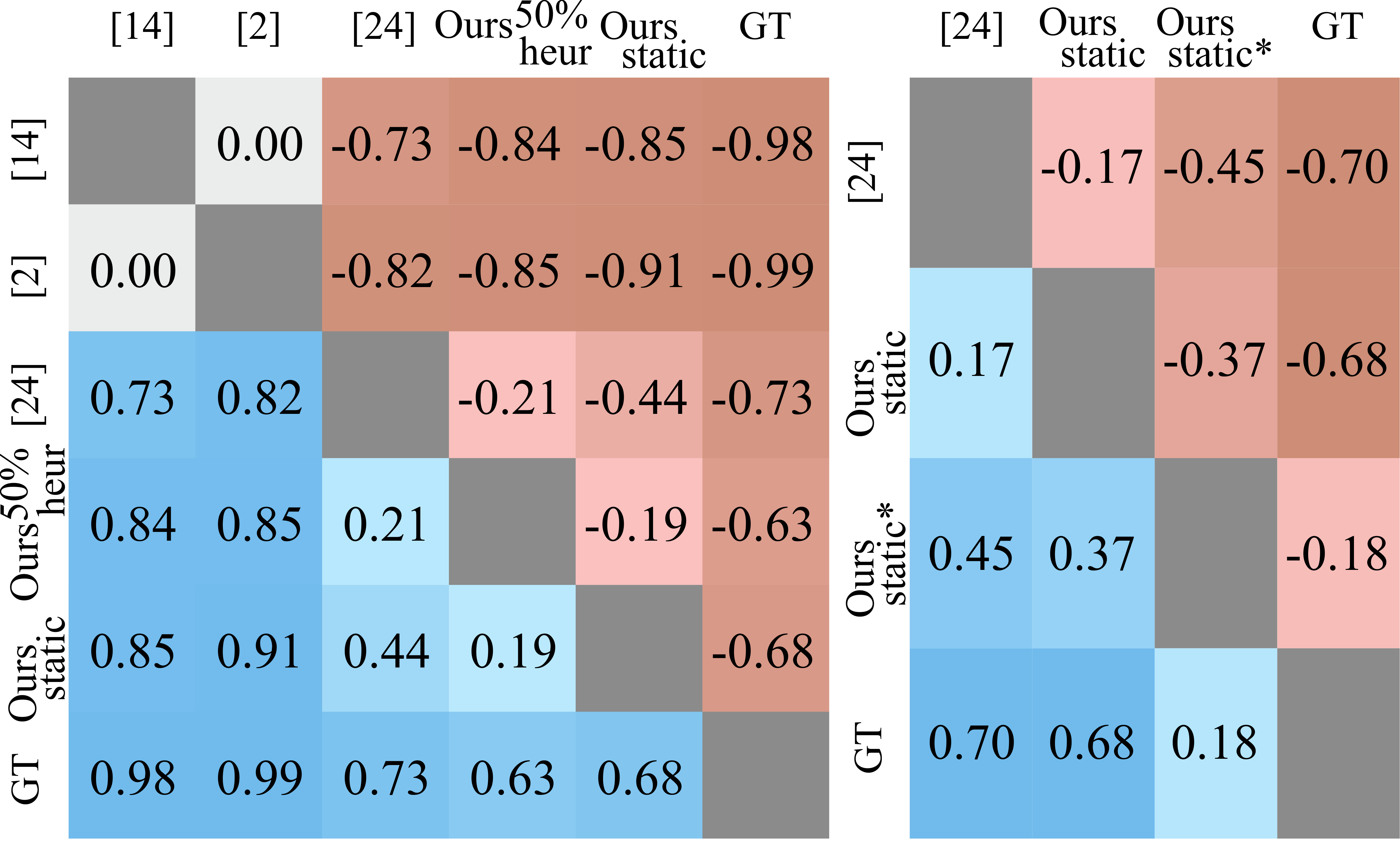}
      \caption{Realism scores based on the user study for each pair of methods (or GT).
      %
      The tables are to be read row-by-row: The bottom row shows that the GT receives positive scores against all the other methods. The left is the evaluation with raw segmentation masks, and the right is with the vector-floorplan images.
      }
    \label{figure:user_study_scores}
\end{figure}

\section{Implementation Details} \label{sect:results}

We have used PyTorch for implementation and a workstation with 
dual Xeon CPUs and dual NVIDIA Titan RTX GPUs. We adopt the same training configuration as in House-GAN except for the batch size (1 instead of 32) and the number of iterations (500k instead of 300k).~\footnote{WGAN-GP~\cite{gulrajani2017improved}, ADAM 
($b1=0.5$, $b2=0.999$), the learning rate ($0.0001$), the number of critics ($1$), and leaky-ReLUs ($\alpha=0.1$) for all non-linearities except for the last one with hyperbolic tangent.}
For the meta scheme optimization, we use the Python library Hyperopt~\cite{bergstra2013making} with Tree-structured Parzen Estimators. In each round, we compute the target metric on 1,000 bubble-diagrams from the training set, and pick the best scheme after 500 rounds.

%
%
%

\mysubsubsection{Testing framework}
To avoid a network from simply copying and pasting layouts, we use the k-fold cross validation from House-GAN~\cite{nauata2020house}, while dividing the samples into four groups based on the number of rooms: (5, 6, 7, 8): When generating layouts with 8 rooms, we use samples from the other three groups for training.
At test time, we randomly pick a GT layout from the test set, use its bubble-diagram to initialize a relational graph, and generate a layout sample. We repeat the process 1,000 times to generate 1,000 samples for the evaluation. 
For the diversity and compatibility evaluations, we compute the mean and the standard deviation of the metrics over five rounds. The same process is used for the layouts with the other room counts.

\mysubsubsection{Realism user study}
We have carried out the user study with 10 amateurs (i.e. engineers and graduate students) and 10 professional architects on two layout representations. The first is the raw output of the systems as pixel-wise segmentation masks. The second is a vector-floorplan representation where we have used an off-the-shelf floorplan vectorization system (Floor-SP~\cite{chen2019floor}) for the conversion of the samples except for the ground-truth.
Each room/door is assigned a unique color based on its type as in Fig.~\ref{figure:groundtruth}. A subject is presented a pair of layouts as in Fig.~\ref{figure:realism}, 
%
and asked which one is more realistic or a tie. A method scores ($+1$) for a win, ($-1$) for a loss, and ($0$) for a tie. For each pair of methods, we generate 100 pairs to be presented to 10 amateurs and 10 professional architects,
where the average score becomes the realism metric. 
Please see the supplementary document for the full details of the user study.


\mysubsubsection{Competing methods}
We compare against the 
three competing methods with their official implementations: 
House-GAN~\cite{nauata2020house}, Ashual \etal~\cite{ashual2019specifying}, and Johnson \etal~\cite{johnson2018image}.
For House-GAN~\cite{nauata2020house}, the room shapes were simplified to rectangles via pre-processing in their work. In our experiments, non-rectangular room shapes are used for fair comparison.
%
For Ashual \etal~\cite{ashual2019specifying} and Johnson \etal~\cite{johnson2018image}, we convert our bubble-diagram and floorplan data into their scene-graph representations, while
limiting to three connection types (``room-room", ``room-door", or ``none"). 




\section{Experimental Results}
Table~\ref{tab:main_results} provides the main results, where we use the proposed approach with a baseline refinement scheme ($\mbox{Ours}^{50\%}_{heur}$).
%
The table shows that our system outperforms all the other methods in all the metrics with clear margins, which are the greatest for the most challenging task (i.e., column ``8'', generating layouts of 8 rooms, where layouts of 5, 6, or 7 rooms are used for training).
Note that a realism metric is first computed for pairs of methods as described above, then their average is reported for each method.
%



%

Qualitative evaluations on realism, diversity, and compatibility are given in Figs.~\ref{figure:realism}, \ref{figure:diversity}, and \ref{figure:compatibility}, respectively.
The figures show that our layouts are more realistic with better spatial arrangement of rooms and their individual shapes. House-GAN does a reasonable job especially in the diversity evaluation, but doors are often missing and room boundaries have noticeable artifacts.
Compatibility is more challenging in our problem, where the functional connectivity is given in the bubble diagram and small doors need to be placed precisely at the room boundaries.
%
%
Fig.~\ref{figure:iterations} demonstrates how the iterative refinement improves the quality of a design over time with our method $\mbox{Ours}_{static}$. Both the room spatial arrangement and their shapes make significant improvements over the iterations.

The complete pairwise realism scores are provided in Fig.~\ref{figure:user_study_scores}. The left is the results with raw segmentation visualization, where two of our methods ($\mbox{Ours}^{50\%}_{heur}$ and $\mbox{Ours}_{static}$) are compared against the three competing methods and the ground-truth.
Our scheme-optimized system ($\mbox{Ours}_{static}$) is the best among the competing methods but is still behind GT with a big margin. In direct comparisons between $\mbox{Ours}_{static}$ and GT, subjects choose ``Ours is better'' only 6\% of the time (``tie'' is 20\%). The right is the results with the vector-floorplan visualization, comparing against the best competing method House-GAN and the ground-truth. For this evaluation, in addition to $\mbox{Ours}_{static}$, we have prepared another variant $\mbox{Ours}_{static*}$ where only fully compatible samples (roughly 10\% in our generations) are used in the evaluation.
While GT is still the winner, $\mbox{Ours}_{static*}$ scores -0.18 against GT, where $0.0$ is the ultimate score when our results become indistinguishable from GT. Subjects chose ``Ours is better'', ``tie'', and ``GT is better'' at 25\%, 32\%, and 43\% of the time, respectively, indicating that even professional architects struggle to distinguish our results from GT (See Figs.~\ref{figure:user_study} and~\ref{figure:user_study_scores}).

Lastly, Table~\ref{tab:ablation} demonstrates the effectiveness of the refinement scheme optimization, where the diversity or the compatibility is the target metric.
The compatibility optimization requires the information of the current design (i.e., dynamic information) and cannot be optimized with $\mbox{Ours}_{static}$.
Overall, optimized refinement schemes outperform heuristic schemes consistently.
The best compatibility measure is achieved when the compatibility metric is the target as expected. Looking at the scheme parameters, we found that in 9 out of 12 node/edge types, the scheme passes the mask-conditions at earlier iterations when being compatible than when being incompatible, which agrees with our intuition. The best diversity measure is achieved when the diversity is the metric. Looking at the parameters of $\mbox{Ours}_{static}$, we found that the scheme passes the mask-conditions in the order of doors, living room, kitchen, bedroom, and bathroom. This was counter-intuitive at first, because one designs doors last. This is due to the fact that doors are small and the effects of the L1-loss are minimal. We found that the doors often move when the input conditions are specified. The rest of the room order makes sense, as it is easy to start with rooms at the core (i.e., living room or kitchen).

\section{Conclusion}
This paper makes a breakthrough in the task of automated house layout generation via a novel generative adversarial layout refinement network, generating vector floorplans often indistinguishable from ground-truth.
%
While we have evaluated the system for an offline layout generation task, computational capability of refining an incomplete design is also effective in incorporating user inputs: Architects can take a design, makes adjustments, and pass back to the system for refinement.
We believe that this paper sets a milestone in the literature of automated house layout generation, even with potentials in the space of interactive layout design.
We will share our code, model, and data.

\vspace{0.5cm}
\mysubsubsection{Acknowledgement}
This research is partially supported by NSERC Discovery Grants, NSERC Discovery Grants Accelerator Supplements, and DND/NSERC Discovery Grant Supplement. We would like to thank architects and students for participating in our user study.


{\small
\bibliographystyle{ieee_fullname}
\bibliography{egbib}
}

\end{document}